\documentclass[twoside,twocolumn,9pt]{article}
\usepackage{extsizes}
\usepackage[super,sort&compress,comma]{natbib} 
\usepackage[version=3]{mhchem}
\usepackage[left=2cm, right=2cm, top=1.785cm, bottom=2.0cm]{geometry}
\usepackage{balance}
\usepackage{mathptmx}
\usepackage{sectsty}
\usepackage{graphicx} 
\usepackage{lastpage}
\usepackage[format=plain,justification=justified,singlelinecheck=false,font={stretch=1.125,small,sf},labelfont=bf,labelsep=space]{caption}
\usepackage{float}
\usepackage{fancyhdr}
\usepackage{fnpos}
\usepackage[english]{babel}
\addto{\captionsenglish}{%
  
}
\usepackage{array}
\usepackage{droidsans}
\usepackage{charter}
\usepackage[T1]{fontenc}
\usepackage[usenames,dvipsnames]{xcolor}
\usepackage{setspace}
\usepackage[compact]{titlesec}
\usepackage{hyperref}

\usepackage{url}
\usepackage{algorithm}
\usepackage{algpseudocode}
\usepackage{amsthm}
\usepackage{amsmath}
\usepackage{amssymb}
\usepackage{mathtools, nccmath}
\usepackage{epstopdf}
\usepackage[switch]{lineno}

\definecolor{cream}{RGB}{222,217,201}

\begin{document}

\pagestyle{fancy}
\thispagestyle{plain}
\fancypagestyle{plain}{
\renewcommand{\headrulewidth}{0pt}
}

\makeFNbottom
\makeatletter
\renewcommand\LARGE{\@setfontsize\LARGE{15pt}{17}}
\renewcommand\Large{\@setfontsize\Large{12pt}{14}}
\renewcommand\large{\@setfontsize\large{10pt}{12}}
\renewcommand\footnotesize{\@setfontsize\footnotesize{7pt}{10}}
\makeatother

\renewcommand{\thefootnote}{\fnsymbol{footnote}}
\renewcommand\footnoterule{\vspace*{1pt}%
\color{cream}\hrule width 3.5in height 0.4pt \color{black}\vspace*{5pt}} 
\setcounter{secnumdepth}{5}

\makeatletter 
\renewcommand\@biblabel[1]{#1}            
\renewcommand\@makefntext[1]%
{\noindent\makebox[0pt][r]{\@thefnmark\,}#1}
\makeatother 
\renewcommand{\figurename}{\small{Fig.}~}
\sectionfont{\sffamily\Large}
\subsectionfont{\normalsize}
\subsubsectionfont{\bf}
\setstretch{1.125} 
\setlength{\skip\footins}{0.8cm}
\setlength{\footnotesep}{0.25cm}
\setlength{\jot}{10pt}
\titlespacing*{\section}{0pt}{4pt}{4pt}
\titlespacing*{\subsection}{0pt}{15pt}{1pt}

\fancyfoot{}
\fancyfoot[LO,RE]{\vspace{-7.1pt}\includegraphics[height=9pt]{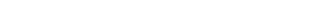}}
\fancyfoot[RO]{\footnotesize{\sffamily{1--\pageref{LastPage} ~\textbar  \hspace{2pt}\thepage}}}
\fancyfoot[LE]{\footnotesize{\sffamily{\thepage~\textbar\hspace{3.45cm} 1--\pageref{LastPage}}}}
\fancyhead{}
\renewcommand{\headrulewidth}{0pt} 
\renewcommand{\footrulewidth}{0pt}
\setlength{\arrayrulewidth}{1pt}
\setlength{\columnsep}{6.5mm}
\setlength\bibsep{1pt}

\makeatletter 
\newlength{\figrulesep} 
\setlength{\figrulesep}{0.5\textfloatsep} 

\newcommand{\topfigrule}{\vspace*{-1pt}%
\noindent{\color{cream}\rule[-\figrulesep]{\columnwidth}{1.5pt}} }

\newcommand{\botfigrule}{\vspace*{-2pt}%
\noindent{\color{cream}\rule[\figrulesep]{\columnwidth}{1.5pt}} }

\newcommand{\dblfigrule}{\vspace*{-1pt}%
\noindent{\color{cream}\rule[-\figrulesep]{\textwidth}{1.5pt}} }

\makeatother

\twocolumn[
  \begin{@twocolumnfalse}

\Huge{\textbf{Real-time High-resolution CO$_2$ Geological Storage Prediction using Nested Fourier Neural Operators$\dag$}} \\

 \noindent\large{Gege Wen,$^{a\ast}$ Zongyi Li,$^{b}$ Qirui Long,$^{a}$
Kamyar Azizzadenesheli,$^{c}$ Anima Anandkumar,$^{b,c}$ Sally M. Benson$^{a}$} \\

\noindent\normalsize{Carbon capture and storage (CCS) plays an essential role in global decarbonization. Scaling up CCS deployment requires accurate and high-resolution modeling of the storage reservoir pressure buildup and the gaseous plume migration. However, such modeling is very challenging at scale due to the high computational costs of existing numerical methods. This challenge leads to significant uncertainties in evaluating storage opportunities, which can delay the pace of large-scale CCS deployment. We introduce Nested Fourier Neural Operator (FNO), a machine-learning framework for high-resolution dynamic 3D CO$_2$ storage modeling at a basin scale. Nested FNO produces forecasts at different refinement levels using a hierarchy of FNOs and speeds up flow prediction nearly 700,000 times compared to existing methods. By learning the solution operator for the family of governing partial differential equations, Nested FNO creates a general-purpose numerical simulator alternative for CO$_2$ storage with diverse reservoir conditions, geological heterogeneity, and injection schemes. Our framework enables unprecedented real-time modeling and probabilistic simulations that can support the scale-up of global CCS deployment.
} \\


 \end{@twocolumnfalse} \vspace{0.6cm}

  ]

\renewcommand*\rmdefault{bch}\normalfont\upshape
\rmfamily
\section*{}
\vspace{-1cm}

\footnotetext{\textit{$^{a}$~Energy Sciences and Engineering, Stanford University, Stanford, 94305, CA, USA}}
\footnotetext{\textit{$^{b}$~Computing and Mathematical Sciences, California Institute of Technology, Pasadena, 91125, CA, USA}}
\footnotetext{\textit{$^{c}$~NVIDIA Corporation, Santa Clara, 95051, CA, USA}}

\footnotetext{\dag~Electronic Supplementary Information (ESI) available: Supplementary Text, Figs. S1 to S2, Tables S1 to S7, and References 1-10. See \url{https://www.rsc.org/suppdata/d2/ee/d2ee04204e/d2ee04204e1.pdf}}




\section*{Introduction}

Carbon capture and storage (CCS) is an important climate change mitigation technology that captures carbon dioxide (CO$_2$) and permanently stores it in subsurface geological formations. It provides a tangible solution for decarbonizing hard-to-mitigate sectors and can generate negative emissions when combined with direct air capture or bioenergy technologies~\cite{pathways2019role, luderer2018residual, fankhauser2022meaning}. Most integrated assessment modeling scenarios identify the large-scale global deployment of CCS as a necessity to achieve net-zero emissions by 2050~\cite{riahi2017shared, cozzi2020world}. However, the current pace of CCS deployment has failed to meet expectations~\cite{reiner2016learning}. One of the critical challenges causing the delay is the uncertainties in evaluating storage prospects and injection capacities~\cite{lane2021uncertain}. Injecting CO$_2$ into geological formations leads to pressure buildup and gaseous plume migration~\cite{NationalAcademiesofSciencesEngineering2018}. Forecasts of these dynamic responses are essential for evaluating CO$_2$ storage capacities and guiding engineering decisions. However, current numerical approaches for simulating the CO$_2$-water multiphase flow are highly computationally expensive. As a result, they are inadequate to provide rigorous computation supports that are urgently needed for accelerating CCS project deployments around the world~\cite{lane2021uncertain}. 

\begin{figure*}[!ht]
    \centering
    \includegraphics[width=\textwidth]{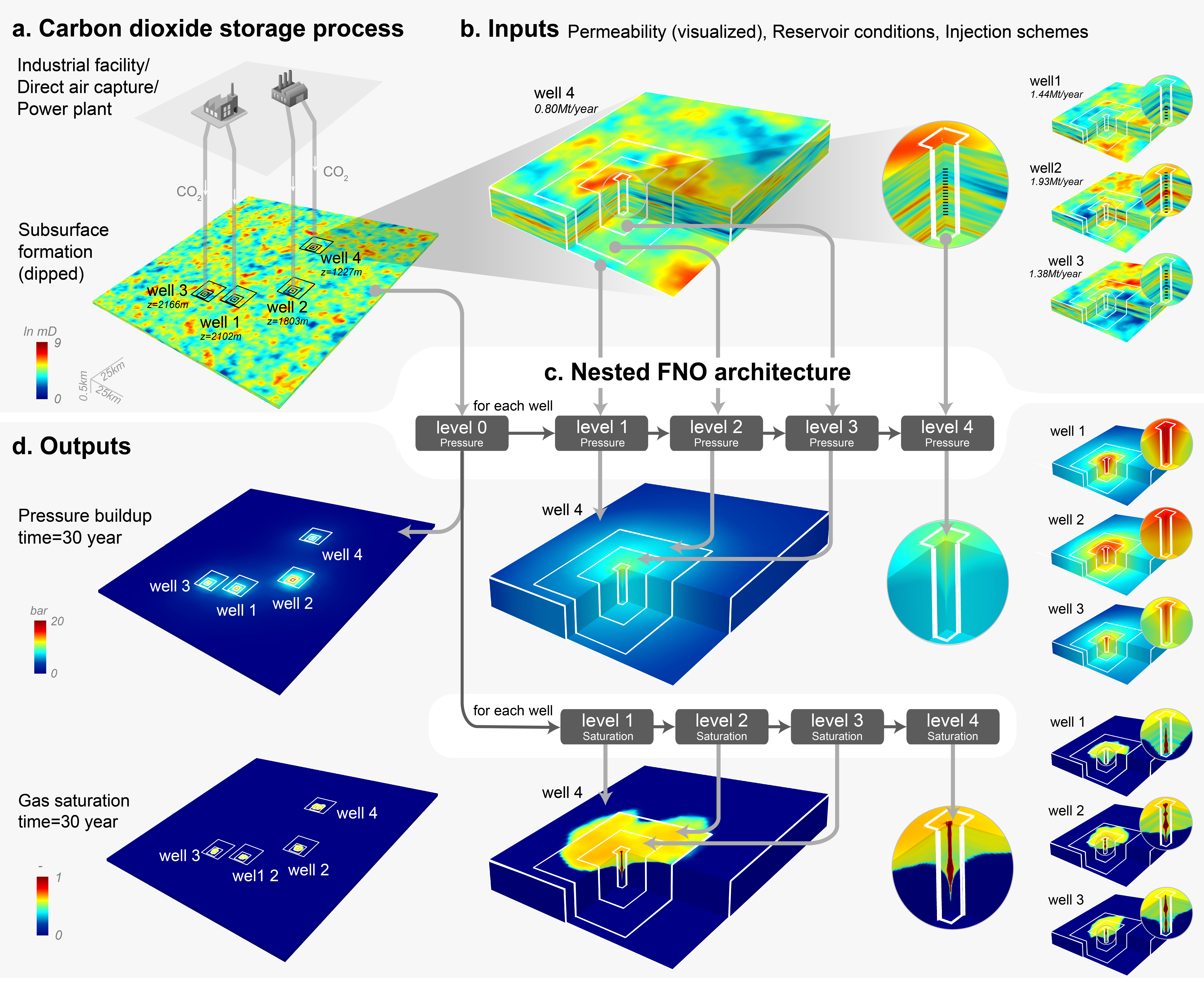}
    \caption{\textbf{Introduction to Nested-FNO.} \textbf{a-b.} Permeability for a dipped 3D reservoir with four injection wells; white and black lines indicate level 0 to 4's boundary; the black dotted lines in the zoomed-in circles show the locations of injection perforation intervals. \textbf{c.} Each grey block represents an FNO model; light grey arrows point to the input and output's level; dark grey arrows show when one model's output is used as another model's input. Notice we feed level 0 pressure buildup output to level 1 gas saturation model because CO$_2$ plumes never migrate to level 0. \textbf{d.} Pressure buildup and gas saturation at 30 years. }
    \label{fig:model_flow}
\end{figure*}

The modeling of CO$_2$ geological storage requires multi-phase~\cite{Pruess1999,blunt2017multiphase}, multi-physics~\cite{pruess2002multiphase}, and multi-scale simulations. The governing partial differential equations (PDEs) involving the multiphase variation of Darcy's law are expensive to solve~\cite{Pruess1999,blunt2017multiphase}. CO$_2$ and water are immiscible and mutually soluble, requiring multi-physics simulation coupled with thermodynamics~\cite{pruess2002multiphase}. 
Moreover, an especially challenging characteristic of CO$_2$ storage modeling is that it demands both high-resolutions and extremely large spatial-temporal domains. 
The  gaseous CO$_2$ plume requires resolutions as fine as one to two meters to provide reliable estimates~\cite{Doughty2010, Wen2019}. Near-well responses such as pressure buildup and the dry-out effect, i.e., evaporation of formation fluid into the gas phase, also require highly resolved grids around the injection well~\cite{pruess2009formation, andre2014well}. Meanwhile, the pressure buildup can travel hundreds of kilometers~\cite{chadwick2004geological} beyond the CO$_2$ plume and interfere with other injection operations. 
Due to these multi-scale responses, many CCS-related analyses are forced to use inaccurate simulations with coarsened grid resolution~\cite{kou2022method} and/or simplified physics~\cite{cavanagh2011simulation}. 

One popular approach for reducing the computational costs of numerical simulations is to use non-uniform grids. Specifically, the local grid refinement (LGR) approach~\cite{bramble1988preconditioning},  has enabled simulations of real-world three-dimensional (3D) CO$_2$ storage projects, where the fine-grid responses capture the plume migration while the coarser grid responses capture the far-field pressure buildup~\cite{eigestad2009geological, faigle2014efficient,kamashev2021reservoir}.
However, even with non-uniform grid approaches, these numerical models are still too expensive to be used for essential tasks such as site selection~\cite{callas2022criteria}, optimization~\cite{nghiem2010simulation,zhang2012numerical}, and inversion~\cite{Strandli2014}, which require probabilistic or repetitive forward stimulations.

In recent years, machine learning approaches are emerging as a promising alternative to numerical simulation for subsurface flow problems~\cite{zhu2018bayesian, mo2019deep, tang2020deep, WEN2021103223, WEN2021104009}. Machine learning models trained with numerical simulation data are usually much faster than simulators, because the inferences of machine learning models are often very cheap. However, standard machine learning methods suffer from the lack of generalization and struggle to provide accurate estimates away from the domain of their training data~\cite{kovachki2021neural}. This limits the usage of machine learning in CO$_2$ storage modeling as it requires generalization under diverse geology, reservoir conditions, and injection schemes. A recent machine-learning framework, named neural operators~\cite{li2020multipole, li2020neural}, overcomes the generalization challenge by directly learning the solution operator for the governing PDE family instead of single instances. Neural operators can generalize to different conditions in the PDE system as well as discretizations without the need for re-training the model. 


Fourier neural operator (FNO) is a type of neural operator that offers especially remarkable predictability for flow-related problems~\cite{li2020fourier}. It uses the Fourier transform to learn the solution operator efficiently. A variant of FNO was recently proposed for predicting 2D CO$_2$-water multiphase flow with great generalization and accuracy~\cite{wen2022u}. However, despite the advances in model generalization, the development of machine learning approaches for CO$_2$ storage is impeded by the multi-scale challenge that both high grid resolution and large spatial domain are required. Previous machine learning models are limited to either 2D problems that can only represent flat reservoirs with a single injection well~\cite{WEN2021104009, wen2022u} or 3D problems with very coarse resolutions that fail to capture essential physics~\cite{tang2022deep,yan2022robust}. In real-world scenarios, CCS projects often involve multiple injection wells and dipped reservoirs. These processes can only be accurately captured by high-resolution 3D simulations, where the costs of collecting training data from numerical simulators become prohibitive. 


Here we present a machine learning framework with an unprecedented capability of high-resolution dynamic 3D modeling for basin-scale CO$_2$ storage. We integrate the FNO machine learning architecture with a semi-adaptive LGR modeling approach for numerical simulation and present the Nested Fourier Neural Operator (Nested FNO) architecture. As shown in Figure~\ref{fig:model_flow}, five levels of FNOs are used to predict flow responses in five different resolutions. This approach vastly reduces the computational cost needed during data collection as well as overcomes the memory constraints in model training. Using this approach, our prediction resolution exceeds many benchmark CO$_2$ storage simulations run with existing numerical models, such as Sleipner benchmark model~\cite{sleipnerdata} and Decatur model~\cite{decaturdata}. Meanwhile, Nested FNO only needs less than 2,500 training data at the coarsest resolution and about 6,000 samples for the finer resolutions. Despite the small training size, it generalizes well to the large problem dimension with millions of cells and a diverse collection of practical input variables. 

In addition, Nested FNO offers \textit{real time} forecasts, where the inference speed is 700,000 times faster compared to the state-of-the-art numerical solver. The fast inference enables many critical tasks for CCS decision-making that were prohibitively expensive. For example, we present a rigorous probabilistic assessment for maximum pressure buildup and CO$_2$ plume footprint. Such assessment can reduce uncertainties in capacity estimation and injection designs~\cite{NationalAcademiesofSciencesEngineering2018}; however, it would have taken nearly two years with numerical simulators. Using Nested FNO, this assessment took only 2.8 seconds. These high-quality real-time predictions can greatly improve our ability to develop safe and effective CCS projects.

\vspace{1em}
\section*{Results \& Discussion}
\subsection*{Data Overview} We consider CO$_2$ injection into 3D saline reservoirs~\cite{page2020global} through multiple wells over 30 years, as shown in Figure~\ref{fig:model_flow}~\textbf{a}. Our data set includes a comprehensive collection of input variables for practical CO$_2$ storage projects, covering most realistic scenarios of potential CCS sites.
Input parameters comprise reservoir conditions (depth, temperature, dip angle), injection schemes (number of injection wells, rates, perforation intervals), and permeability heterogeneity (mean, standard deviation, correlation lengths).  This comprehensive data set allows the trained Nested FNO to serve as a general-purpose simulator alternative for most CCS project settings. 
See \textit{Supplementary Text, Data set generation} for details on input variable sampling. 

The numerical simulation data is generated using a semi-adaptive LGR approach to ensure high fidelity and computational tractability. We use global (level 0) resolution grids in the large spatial domain to mimic typical saline storage formations with infinite boundary conditions. Next, we apply four levels of local refinements (levels 1 to 4) around each well to gradually increase the grid resolutions. Going from levels 0 to 4, we reduce the cell size by 80x on the $x,y$ dimensions and 10x on the $z$ dimension to resolve near-well plume migration, dry-out, and pressure buildup. See \textit{Supplementary Text} for full details on the LGR design, governing PDEs, and numerical simulation setups.

\subsection*{Nested FNO Architecture} 

The computational domain of the Nested FNO is a 3D space with time, \(D = \Omega \times T\), where $T$ is the time interval of 30 years and $\Omega$ is the reservoir domain. We use a sequence of FNO models to predict the 3D reservoir domain consisting of subdomains $\Omega_i$ at levels 0 to 4 (Figure~\ref{fig:model_flow}). At each refinement level, we extend the original FNO~\cite{li2020fourier} architecture into 4D to produce outputs for pressure buildup ($P$) and gas saturation ($S$) in the 3D space-time domain.
See \textit{Supplementary Text, Fourier Neural Operator} for detailed architecture and parameters.

The input for each model includes the permeability field, initial hydro-static pressure, reservoir temperature, injection scheme, as well as spatial and temporal encoding. In CO$_2$ storage, pressure buildup travels significantly faster than gas saturation. Therefore, as shown in Figure~\ref{fig:model_flow}~\textbf{c}, we first use an FNO model to predict the pressure buildup at level 0 to capture the global propagation as well as the interaction between wells. We then feed level 0 pressure buildup predictions around each injection well ($\hat{P}_{\Omega_0}|_{wellj}$) to the FNO models on level 1. Each subsequent model takes the input on domain $\Omega_{i}$ together with the coarser-level prediction of $\hat{P}$ or $\hat{S}$ on \(\Omega_{i-1}\), and outputs the predictions of $\hat{P}$ or $\hat{S}$ on \(\Omega_{i}\). By giving the coarser-level prediction to the finer-level model as an input, we also provide the boundary conditions of the finer-level subdomain, which significantly improves the finer-level predictions. 

\subsection*{CO$_2$ Plume Predictions}
\begin{figure*}[t]
    \centering
    \includegraphics[width=\textwidth]{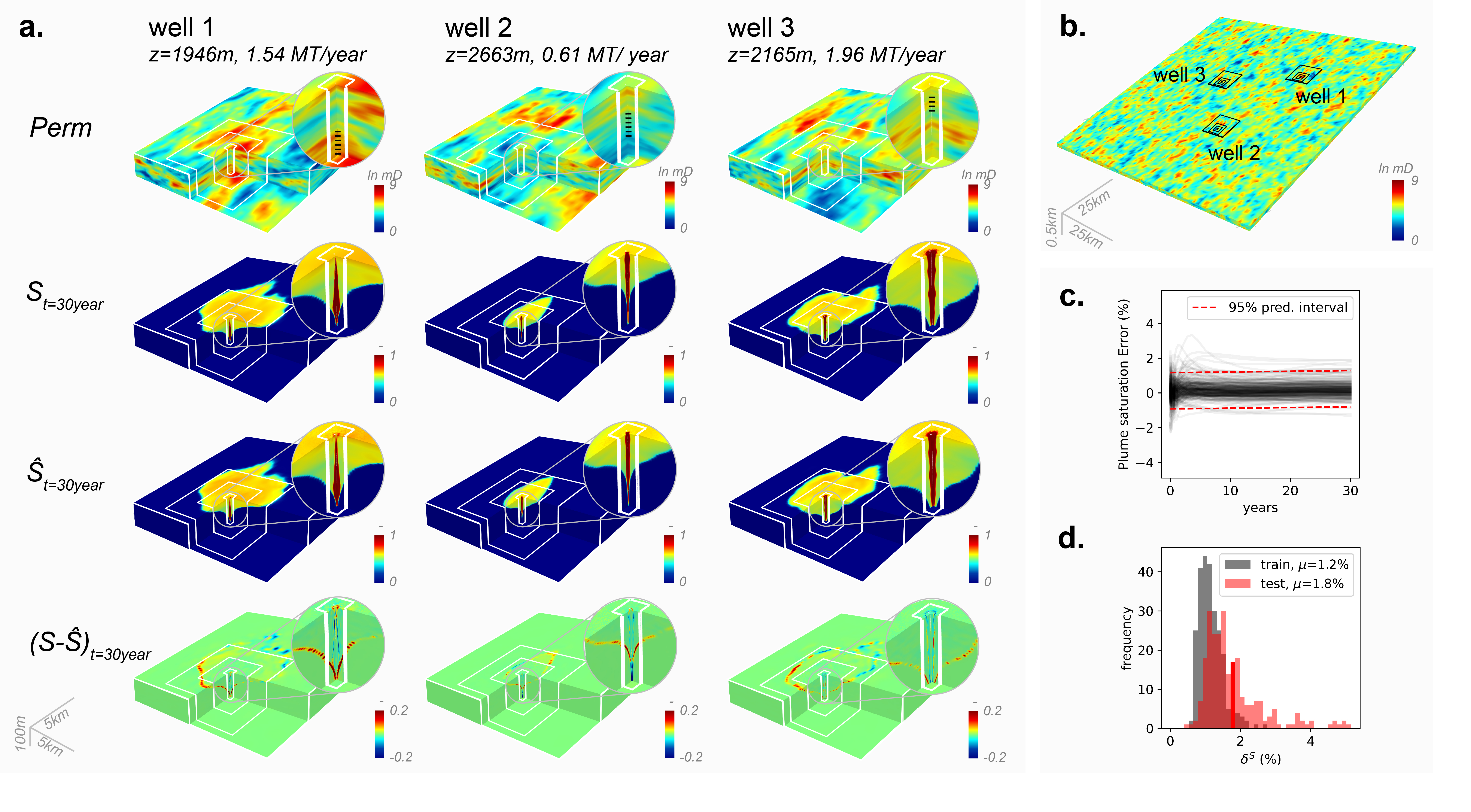}
    \caption{\textbf{Gas saturation prediction.} \textbf{a.} Visualizations of gas saturation predictions at 30 years for a 3-well case. Each row shows permeability, gas saturation ground truth, prediction, and error. The white lines indicate the boundary between each level. \textbf{b.} Reservoir permeability and the location of each well. \textbf{c.} Testing set plume saturation error versus time for 250 random cases. The red dotted line shows the 95\% prediction bands of the error. \textbf{d.} Error histograms for 250 cases in the training and test set. The solid red column indicates the error for the example shown in a. }
    \label{fig:accuracy}
\end{figure*}

The migration of CO$_2$ plume is governed by the complex interplay of viscous, capillary, and gravity forces. As shown in Figure~\ref{fig:model_flow}~\textbf{a}, CO$_2$ plumes tend to migrate up-dip due to buoyancy. They can also form distinctively different shapes and sizes according to different injection rates, perforation intervals, and permeability heterogeneity. The reservoir conditions, i.e., initial hydro-static pressure and temperature, determine CO$_2$ and water fluid properties, which also influence the plume migration. Due to the presence of the dip angle, reservoir conditions can vary significantly even in the same basin (Figure~\ref{fig:model_flow}~\textbf{a}). 

As shown in Figure~\ref{fig:accuracy}~\textbf{a}, Nested FNO successfully captures all of these complex processes. The shapes and saturation distribution of each plume are accurately predicted for each well. Near the injection perforation interval, we observe dry-out zones where the gas saturation is almost one. Dry-out may cause the precipitation of salt in saline formations and can lead to potential loss of permeability and injectivity~\cite{pruess2009formation, miri2015new}. Such processes are neglected by many numerical models as they require high grid resolutions and full physics simulations. Nevertheless, Nested FNO predicts the dry-out zone with excellent accuracy. We observe slightly more errors at the edges of gaseous plumes and dry-out zones; this is because the numerical simulation training data around discontinuous saturation transitions consist of inherent numerical artifacts that are less systematic.

Overall, Nested FNO displays great generalization with small overfitting (Figure~\ref{fig:accuracy}~\textbf{d}). The average  saturation error ($\delta_S$) for the gaseous CO$_2$ plume  is 1.2\% for the training set and 1.8\% for the testing set. See \textit{Experimental} for the definition of the evaluation metric $\delta_S$. This accuracy is well sufficient for most practical applications, such as estimating sweep efficiencies as well as forecasting plume footprints for land acquisition or monitoring program design.

\subsection*{Near-well and Far-field Pressure Buildup Predictions}
\begin{figure*}[t]
    \centering
    \includegraphics[width=\textwidth]{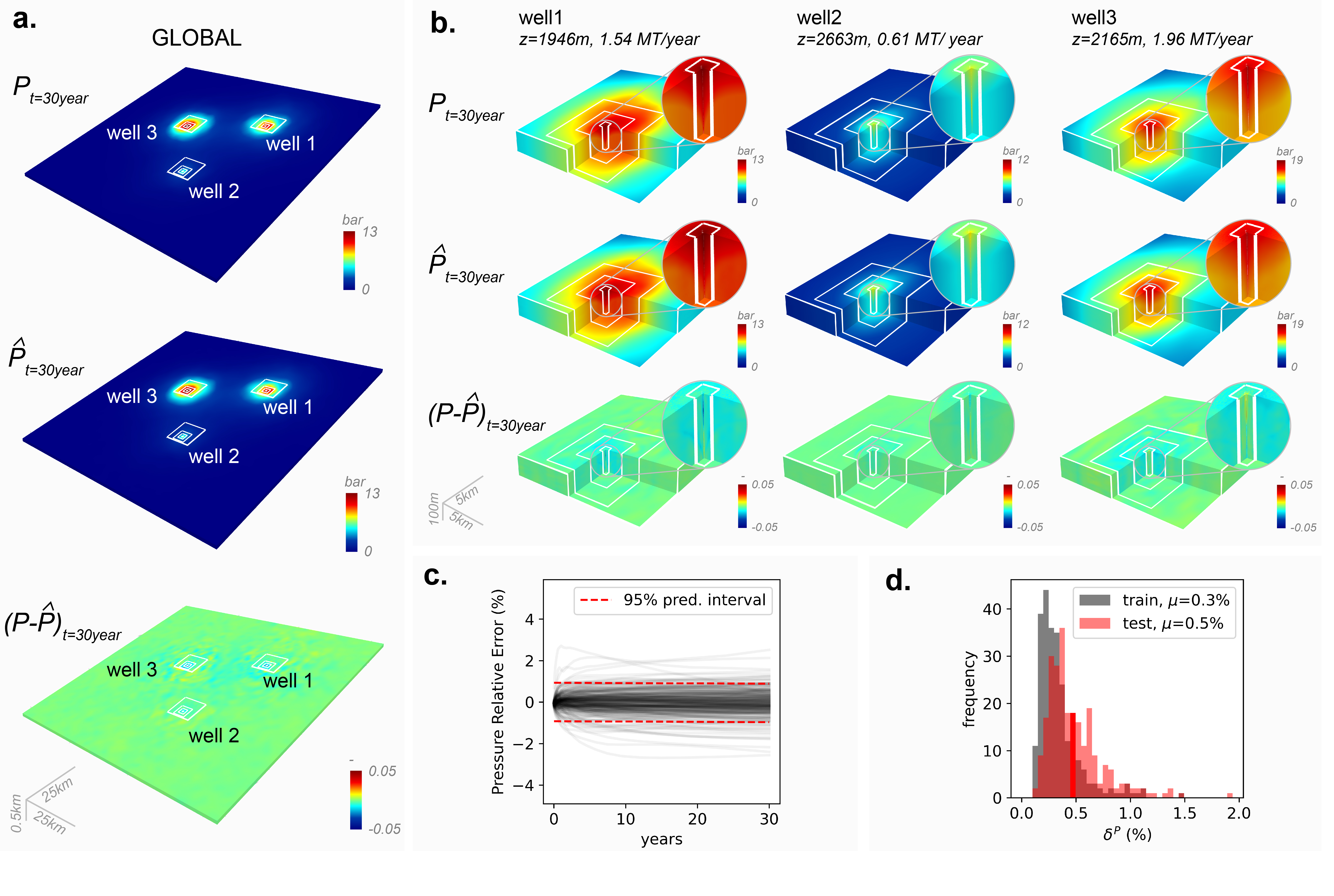}
    \caption{\textbf{Pressure buildup prediction.} \textbf{a. } Global and \textbf{b. }well pressure buildup predictions at 30 years. Each row shows pressure buildup ground truth, prediction, and relative error. The white lines indicate the boundary between each level. \textbf{c.} Testing set pressure relative error versus time for 250 random cases. The red dotted line shows the 95\% prediction bands of the error. \textbf{d.}  Error histograms for 250 cases in the training and test set. The solid red column indicates the error for the visualized example. }
    \label{fig:dp_accuracy}
\end{figure*}

For basin-scale CCS projects, pressure buildups caused by different injection activities can interfere with one another.
As demonstrated in Figure~\ref{fig:dp_accuracy}, Nested FNO precisely captures the local pressure buildup responses around each well, as well as the global interaction among them. The high resolution refinements provide accurate estimates of the maximum pressure buildup, which is an essential indicator of reservoir integrity. The global level prediction provides the spatial extent of the region of pressure buildup influence, another important parameter required for regulatory purposes~\cite{epa}. Additionally, Figure~\ref{fig:dp_accuracy}~\textbf{a-c} shows that  accuracy is consistent across different resolutions and  throughout the injection period. These predictions are sufficient to guide important engineering decisions, such as choosing injection rates. 

The relative pressure buildup error $\delta^{P}$ (as defined in \textit{Experimental}) for the training and the testing set are 0.3\% and 0.5\%, respectively. Similar to the gas saturation, we observe small overfitting from the error histogram (Figure~\ref{fig:dp_accuracy}~\textbf{d}) for the training and testing set. This generalization is remarkable, considering the small training data size for this high-dimensional problem. The generalizability is achieved through a novel fine-tuning technique which we discuss in \textit{Experimental}.

\subsection*{Computational Speed-up}
Once Nested FNO is trained, the diverse input range allows it to act as a computationally efficient alternative to simulators. Users can skip traditional simulations and directly obtain high-fidelity and high-resolution predictions by inferring the trained machine learning models~\cite{WEN2021104009}. Our approach differs from the task-specific ``surrogate'' modeling approach~\cite{zhu2018bayesian, tang2020deep, tang2021deep}, which only considers a specific set of reservoirs for a certain use case. 

We analyze the computational speedup by comparing the Nested FNO's prediction time to the numerical simulation run time of a state-of-the-art full-physics simulator ECLIPSE~\cite{eclipse}. Nested FNO's prediction time varies from 0.025s to 0.085s depending on the number of injection wells. On average, the Nested FNO provides 400,000 (1-well case) to  700,000 (4-well case) times speedup compared to ECLIPSE. Refer to \textit{Supplementary Text, Speedup analysis} for detailed specifications for each method.

\subsection*{Probabilistic Assessment}

\begin{figure*}[t]
    \centering
    \includegraphics[width=\textwidth]{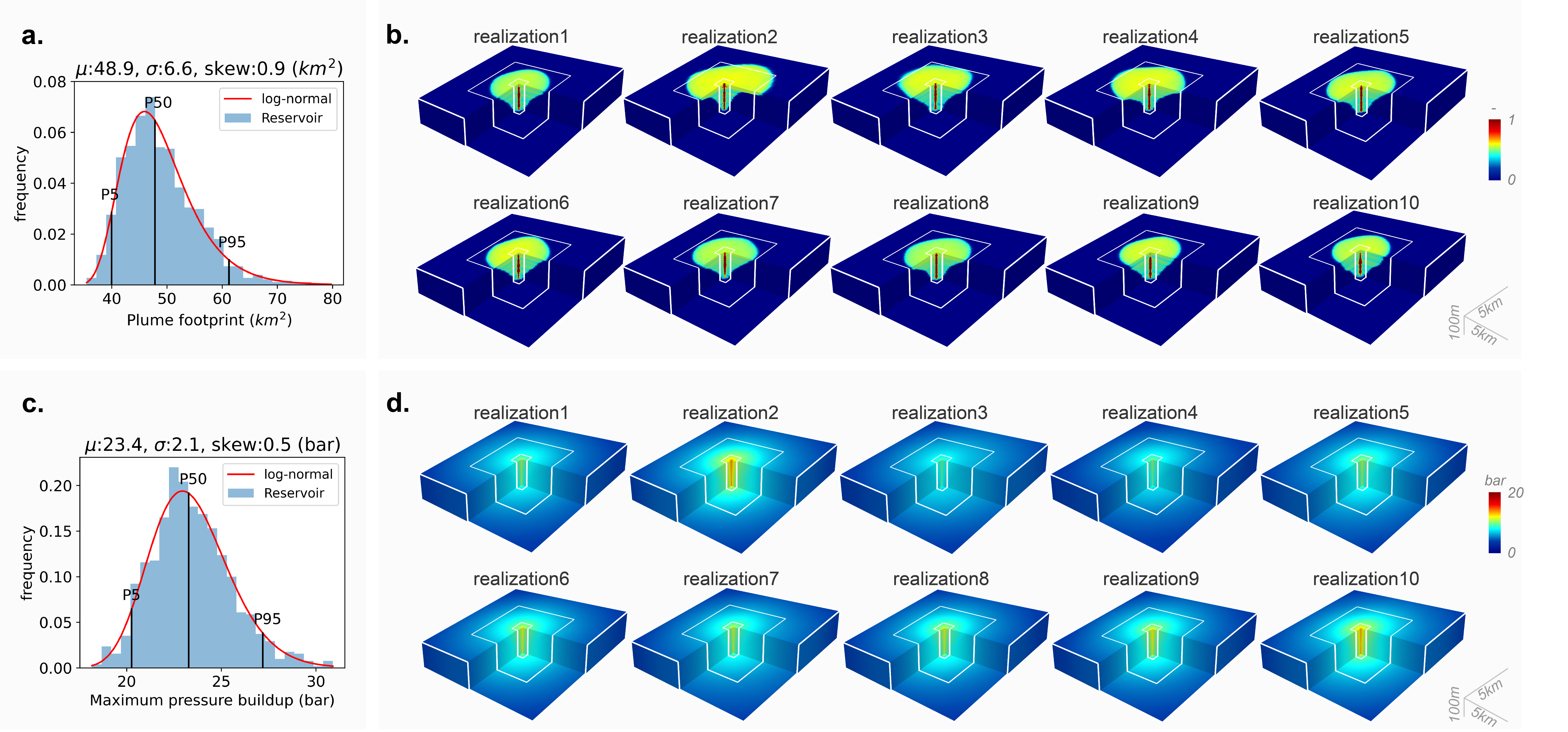}
    \caption{\textbf{Probabilistic assessment.} \textbf{a.} Histogram of CO$_2$ plume footprint predictions given 1,000 permeability realizations from the same geological parameters. The result satisfies a log-normal  distribution; P5, P50, and P95 are marked on the distribution. \textbf{b.} Ten realizations of CO$_2$ plume at 30 years. \textbf{c.} Histogram of CO$_2$ pressure buildup predictions given the same 1,000 permeability realizations. The result satisfies a log-normal distribution; P5, P50, and P95 are marked on the distribution. \textbf{d.} Ten realizations of pressure buildup at 30 years. }
    \label{fig:prob}
\end{figure*}

Nested FNO's fast prediction speed enables rigorous ensemble modeling and probabilistic assessments that were previously unattainable. As an example, we conducted a probabilistic assessment for the maximum pressure buildup and CO$_2$ plume footprint for a four-well CCS project where each well injects at a 1MT/year rate. To investigate the influence of permeability heterogeneity, we generate 1,000 realizations using a fixed set of distribution and spatial correlations, then use Nested FNO to predict gas saturation plumes and pressure buildup for each realization. Refer to \textit{Supplementary Text, Probabilistic assessment} for detailed setups. 
As shown in Figure~\ref{fig:prob}, we obtained probabilistic estimates of the CO$_2$ plume footprint and maximum pressure buildup, which can help project developers and regulators manage uncertainties~\cite{pawar2016national}. For example, the plume footprint helps determine the area of the land lease acquisition required~\cite{U.S.DepartmentofEnergy2017}; the maximum reservoir pressure buildup helps evaluating the safety of a certain injection scheme and ensures reservoir integrity. Running this assessment takes only 2.8 seconds with Nested FNO but requires nearly two years with traditional numerical simulators.

\subsection*{Generalizability of FNO-typed Architectures}

A highlight of the Nested FNO is its excellent generalizability. The training sizes for Nested FNO are small (2,408 for the coarsest model and 5,916 for finer models), considering the large problem dimension with millions of cells. We achieve this generalizability through (1) a novel fine-tuning technique for the nested architecture 
as introduced in \textit{Experimental} and (2) the utilization of the FNO architecture. Most existing data-driven machine learning approaches for subsurface flow use a convolutional neural network (CNN)-based architecture. CNN models' local kernels and deep architectures make them prone to overfitting~\cite{jiang2021deep, wu2020physics, kadeethum2021framework,tangmeng2021deep,tang2021deep}. 
Unlike CNN, FNO uses global kernels to learn an infinite-dimensional input-output mapping in the function space~\cite{li2020fourier}. As a result, using FNO greatly reduces the demand for training data; combining the FNO architecture with the semi-adaptive LGR approach makes this high-resolution dynamic 3D problem tractable. 

Besides data-driven approaches, another line of work, often referred to as a physics-informed neural network, attempts to solve the governing PDE by parameterizing governing relations and initial/boundary conditions using neural networks~\cite{raissi2019physics}. However, these approaches have not yet shown significant advantages in computational efficiency for multiphase flow problems with heterogeneous media~\cite{fuks2020physics, almajid2021prediction, fraces2021physics,haghighat2022physics}. On the contrary, Nested FNO demonstrates the great potential of data-driven approaches not only for CO$_2$ storage but also for other environmental and energy problems that involve multi-scale modeling. For example, in weather forecast modeling, different cyclones can develop locally while interacting with each other on a global level~\cite{dong1983relative}; in nuclear fusion, the collision of multiple nuclei in a particle involves long-distance interactions as well as inner-nucleus many-particle physics~\cite{jin2021lise}.

\section*{Conclusions}
We present Nested FNO for predicting high-resolution dynamic 3D gas saturation and pressure buildup in CO$_2$ storage problems. The trained model provides exceptionally fast predictions and can support many tasks in the CCS deployment that require repetitive forward simulations, including but not limited to (1) probabilistic assessment - as demonstrated above, (2) site selection~\cite{callas2022criteria} - quick screening for a large number of potential reservoirs, (3) storage optimizations~\cite{kumar2008optimizing,zhang2012numerical} - exhaustive search in the parameter space, and (4) seismic inversion~\cite{yin2022learned} - provide simulation outputs and gradients. Nested FNO can facilitate rigorous analyses for these tasks, therefore help reducing uncertainties and accelerate the CCS deployment scale-up progress. 

In addition to providing fast and accurate forecasts, Nested FNO also promotes equity in CO$_2$ storage project development and knowledge adoption. This especially benefits small- to mid-sized developers~\cite{global2016understanding} as well as communities that desire independent evaluation of projects being proposed. High-quality forecasts were previously unattainable for these important players.

\section*{Experimental}

\subsection*{Training procedure}\label{training_procedure_main}
The Nested FNO architecture consists of models $\mathcal{G}^P_{0\ldots 4}$ for pressure buildup and $\mathcal{G}^S_{1\ldots 4}$ for gas saturation. To train these models, we first prepare the input-output pairs for each subdomain and train each of the nine models independently. For each model, we use the ground truth numerical simulation pressure buildup and gas saturation on the coarse-level training domain to construct the input. This approach is time efficient because it allows us to train all models concurrently instead of sequentially going from coarser-level to finer-level models. Refer to \textit{Supplementary Text, Training procedure} for full details. 

\subsection*{Inference procedure}
\begin{algorithm}[t]
    \caption{Predict gas saturation and pressure buildup in a reservoir with $n$ injection wells. $\mathcal{G}$ denotes the a model, $P$ denotes pressure buildup, $S$ denotes gas situation, and $a$ denotes input. }
	\begin{algorithmic}
	\State Use $\mathcal{G}^P_{0}$ to predict $\hat{P}_{0}$  given $a_{0}$
	\For {each well $j=1,\ldots,n$}
	    \State Construct input $(a_{1,j}$, $\hat{P}_{0}|_{j})$
		\State Use $\mathcal{G}^{P}_{1}$ and above input to predict $\hat{P}_{1,j}$
		\State Use $\mathcal{G}^{S}_{1}$  and above input to predict $\hat{S}_{1,j}$ 
		\For {each level $i=2,\ldots,4$}
		    \State Construct input $(a_{i,j}$, $\hat{S}_{i-1,j})$ 
			\State Use $\mathcal{G}^S_{i}$  and above input to predict $\hat{S}_{i,j}$ 
			\State Construct input $(a_{i,j}$, $\hat{P}_{i-1,j})$ 
			\State Use $\mathcal{G}^P_{i}$  and above input to predict $\hat{P}_{i,j}$ 
        \EndFor
	\EndFor
	\end{algorithmic} \label{algorithm}
\end{algorithm}

Once we train the nine models in the Nested FNO, we can predict the gas saturation and pressure buildup according to Algorithm~\ref{algorithm}. The inference input can be constructed given any random combination of reservoir condition (depth, temperature, and dip angle), injection scheme (number of wells, rate, location, perforation interval), and permeability field, as long as the variables are within the training data sampling ranges. Notice that the number of subdomains in $\Omega$ depends on the number of injection wells. For example, a reservoir with three injection wells has 13 subdomains $\Omega=\{\Omega_{0}, \Omega_{level1...4,well1}, \Omega_{level1...4,well2}, \Omega_{level1...4,well3}\}$. We repeat the inference procedure for each injection well.

\subsection*{Evaluation metrics}
To evaluate the gas saturation prediction accuracy in reservoirs with multiple levels of refinements, we introduce the plume saturation error $\delta_S$, defined as:

\begin{align}
    \delta^{S} &= \frac{1}{\sum{I_{t,i}}} \sum_{t\in T} \sum_{i\in \Omega}  I_{t,i} |S_{t,i}-\hat{S}_{t,i}|,  \nonumber \\
    I_{t,i} &= 1 \quad  if  \quad  (S_{t,i}> 0.01) \cup  (|\hat{S}_{t,i}| >
    0.01).
\end{align}
$S$ is the ground truth gas saturation, $\hat{S}$ is the predicted gas saturation, $T$ includes all times snapshots over the 30 years, and $\Omega$ includes all the cells as in the original domain of the numerical simulator; refer \textit{Supplementary Text, Training procedure} for more details. We use this metric because the reservoir domain includes many cells with zero gas saturation; taking an average with these zero predictions leads to an overestimation of the gas saturation accuracy. $\delta_S$ is a more strict metric focusing on the error within the plume. 

For pressure buildup, we introduce relative error $\delta^{P}$:
\begin{align}
    \delta^{P} &= \frac{1}{n_{\Omega}n_{T}} \sum_{t\in T} \sum_{i\in \Omega}  \frac{|P_{t,i}-\hat{P}_{t,i}|}{P_{t,max}}
\end{align}
Here $P$ is the ground truth pressure buildup given by numerical simulation, $\hat{P}$ is the predicted pressure buildup, $P_{t,max}$ is the maximum reservoir pressure buildup at time $t$, $n_\Omega$ is the number of cells in $\Omega$, and $n_T$ is the number time steps. The relative error metric is commonly used for evaluating reservoir pressure buildup~\cite{tang2021deep, wen2022u}.

\begin{figure*}[th]
    \includegraphics[width=0.9\textwidth]{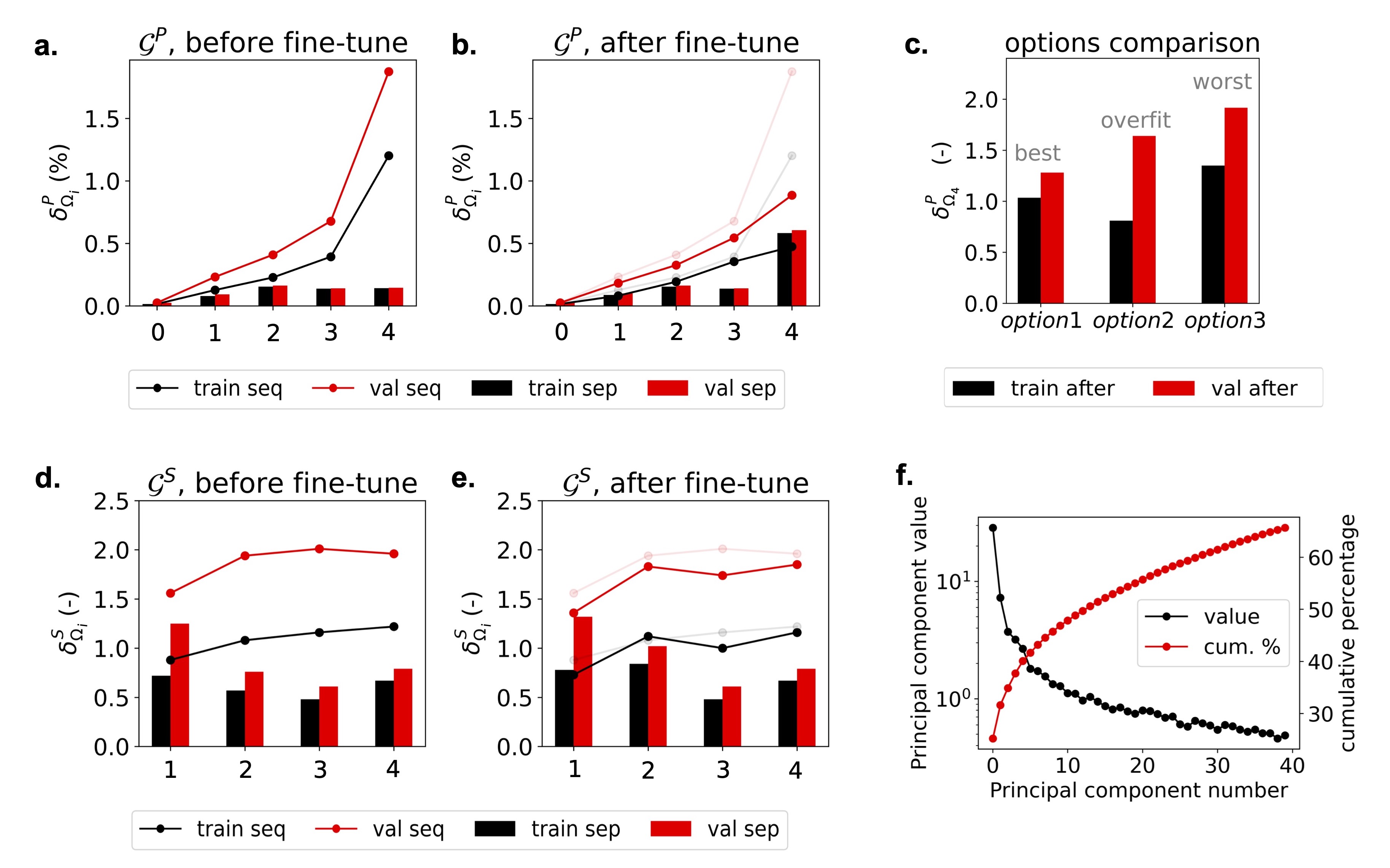}
    \caption{\textbf{Fine-tuning.} Each model's separate and sequential error for \textbf{a.} pressure buildup before fine-tuning, \textbf{b.}  pressure buildup  after fine-tuning, \textbf{d.} gas saturation before fine-tuning, and \textbf{e.} gas saturation after fine-tuning. On the legend, `seq' denotes sequential prediction, `sep' denotes separate prediction. The transparent lines indicate the before fine-tune error. \textbf{e.} Training and validation set $\delta^P_{4}$ of fine-tuning using Option 1 to 3. \textbf{f.} Principle component number and cumulative percentage of the 40 strongest rank for $\mathcal{G}^P_{3}$'s error. }
    \label{fig:finetune}
\end{figure*}

\subsection*{Fine-tuning procedure}
\textbf{Separate vs. sequential prediction.} As described in Algorithm~\ref{algorithm}, during inference, the input for each model in levels 1 to 4 consists of $\hat{S}$ or $\hat{P}$ predicted by their corresponding coarser-level model. However, during training, the inputs are constructed by ground truth numerical simulation data. The discrepancy in training and inference leads to error accumulation, especially for the models that appear later in the prediction sequence. 

To investigate this effect, we introduce two ways to evaluate each model: (1) separate prediction using the ground truth input taken from the numerical simulation (as in training), and (2) sequential prediction using predicted values from the coarser level as input (as in inference). Figure~\ref{fig:finetune}~\textbf{a} compares the average relative pressure buildup $\delta^{P}_{\Omega_j}$ for each model using both separate and sequential prediction methods. Unlike $\delta^{P}$, $\delta^{P}_{\Omega_j}$ focuses on the ability of each model to produce outputs similar to the training data, defined as:

\begin{align}
    \delta^{P}_{\Omega_j} = \frac{1}{n_T n_{\Omega_j}} \sum_{t\in T} \sum_{i\in \Omega_j} \frac{|P_{t,i}-\hat{P}_{t,i}|}{P_{t,max}}.
\end{align}

\noindent Figure~\ref{fig:finetune}~\textbf{a} shows that all models have low errors and negligible overfitting when using separate predictions. However, with sequential prediction, $\delta^{P}_{\Omega_j}$ quickly accumulates, going from coarser to finer-level models. The validation error of level 4 using sequential prediction increased by 13 times compared to separate predictions. 

Similarly, for the gas saturation, the plume gas saturation error $\delta^{S}_{\Omega_j}$ for each model is defined as:

\begin{align}
    \delta^{S}_{\Omega_j} &= \frac{1}{\sum{I_{t,i}}} \sum_{t\in T} \sum_{i\in \Omega_{j}}|S_{t,i}-\hat{S}_{t,i}|
    \nonumber \\
    I_{t,i} &= 1 \quad  if  \quad  (S_{t,i}> 0.01) \cup  (\hat{|S_{t,i}|} > 0.01)
\end{align}

\noindent Figure~\ref{fig:finetune}~\textbf{d} compares $\delta^{S}_{\Omega_j}$ using separate verses sequential prediction. We observed less error accumulation for gas saturation than pressure buildup, which indicates that the prediction of gas saturation does not rely as heavily on coarser-level models.

\vspace{0.5cm}
\noindent\textbf{Random perturbation.}\label{sec:fint_tune}
To reduce the error accumulation, we explored several fine-tuning techniques to improve generalizability using the level 4 pressure prediction as an example. To fine-tune $\mathcal{G}^P_{4}$, we add a perturbation to the ground truth input, $P'_{3,i} = P_{3,i}+\zeta_i$ where $i$ represents a sample taken from the training set. We defined the coarser-level model's error in the training set as $\epsilon_{3}=\hat{P}_{3}-P_{3}$, and explore three configurations of perturbation $\zeta_i$.

\begin{itemize}
    \item Option 1: $\zeta_i = \epsilon_{3,j}$ - randomly sample an instance from $\epsilon_{3}$.
    \item Option 2: $\zeta_i = \epsilon_{3,i}$ - choose the error corresponding to the specific training sample (i.e., fine-tune with the predicted label $\hat{P}_{3,i}$).
    \item Option 3: $\zeta_i\sim \mathcal{N}(\mu_{\epsilon_{3}}, \sigma_{\epsilon_{3}})$ - generate a random Gaussian error using the mean and standard deviation of $\epsilon_{3}$.
\end{itemize}

As shown in Figure~\ref{fig:finetune}~\textbf{c}, Option 1 provides the best validation set performance with the smallest overfitting. By providing a randomly sampled noise instance from $\epsilon_{3}$ with each training data, we let the finer-level models become aware of the presence of a structured error and learn to filter it out. Option 2 gives the best training set error but is significantly overfitted. Interestingly, Option 3 leads to the largest errors in both the training and validation set despite being a commonly used machine learning technique for introducing randomness. 

As a result of this experiment, we applied Option 1 to $\mathcal{G}^P_{1}$, $\mathcal{G}^P_{4}$, $\mathcal{G}^S_{1}$, and $\mathcal{G}^S_{2}$. After fine-tuning, sequential prediction errors are reduced for both pressure buildup and gas saturation (Figure~\ref{fig:finetune}~\textbf{b} and ~\textbf{e}). For pressure buildup, we observe a dramatic improvement on level 4, where the validation error decreased by more than 50\%.

\vspace{0.5cm}
\noindent\textbf{Structure of prediction error.} We hypothesize that the FNO model's predictions and errors lie in a perturbed low-dimensional manifold in the output function spaces due to its structure. To verify our hypothesis, we analyzed the functional principle components on error $\epsilon_{3}$~\cite{blanchard2007statistical}. As shown in Figure~\ref{fig:finetune}~\textbf{f}, only a few principal components are needed to describe nearly a third of the error. Gaussian noised functions did not improve prediction because the Gaussian noise resides in an infinite-dimensional space, whereas the actual error only lives in a small linear sub-space.  

\section*{Web application}
The trained Nested FNO models are hosted in web application \href{https:/ccsnet.ai}{CCSNet.ai} to provide real-time predictions upon the publication of this manuscript. Please also see this link for a demonstration of publicly accessible web application for our previous works.

\section*{Data and Code Availability}
The python code for the Nested FNO model architecture and the data set used in training will be available at \href{https://github.com/gegewen/nestfno}{https://github.com/gegewen/nestfno} upon the publication of this manuscript.

\section*{Author Contributions}
G.W. Conceptualization, Methodology, Software, Data acquisition, Data curation, Formal analysis, Investigation, Validation, Visualization, Writing – original draft, Writing – review \& editing. Z.L. Methodology, Investigation, Validation, Writing – original draft, Writing – review \& editing. Q.L. Data acquisition. K.A. Methodology, Software, Investigation, Validation, Writing – review \& editing. A.A. Funding acquisition, Supervision, Writing – review \& editing. S.B. Conceptualization, Formal analysis, Funding acquisition, Methodology, Resources, Supervision, Writing – review \& editing.

\section*{Conflicts of interest}
There are no conflicts to declare.

\section*{Acknowledgements}
The authors gratefully acknowledge Yanhua Yuan from ExxonMobil for many helpful conversations and suggestions. G.W. and S.B. gratefully acknowledge the support by ExxonMobil through the Strategic Energy Alliance at Stanford University and the Stanford Center for Carbon Storage. Z.L. gratefully acknowledges the financial support from the Kortschak Scholars, PIMCO Fellows, and Amazon AI4Science Fellows programs. A.A. is supported in part by Bren endowed chair. 





\bibliography{rsc} 
\bibliographystyle{rsc} 

\end{document}